\documentclass[10pt,twocolumn,letterpaper]{article}

\usepackage{wacv}              

\usepackage{booktabs}       
\usepackage{amsfonts}       
\usepackage{nicefrac}       
\usepackage{microtype}      
\usepackage{xcolor}         
\definecolor{boxcolor}{HTML}{B85450}

\newcommand*\colourcheck[1]{%
  \expandafter\newcommand\csname #1check\endcsname{\textcolor{#1}{\ding{52}}}%
}
\newcommand*\colourcross[1]{%
  \expandafter\newcommand\csname #1check\endcsname{\textcolor{#1}{\ding{55}}}%
}

\newcommand{\cross}{\textcolor{red}{\ding{55}}}

\usepackage{booktabs}
\usepackage{multirow}
\usepackage{algorithm}
\usepackage{algpseudocode}
\usepackage{comment}
\usepackage{caption} 
\usepackage{arydshln}
\usepackage{tikz}       
\usepackage{adjustbox}  
\usepackage{circledsteps}
\usepackage{subcaption}
\usepackage{mathtools}
\usepackage{amsmath}
\usepackage{wrapfig}
\usepackage{colortbl}
\usepackage{xcolor}
\usepackage{tabularray}
\usepackage{nicematrix}
\usepackage{amssymb}
\usepackage{bbm}
\usepackage{url}
\usepackage{enumitem}
\usepackage{lipsum}
\usepackage{svg}
\usepackage{enumitem}
\usepackage{calc}
\usepackage{xcolor,pifont}
\usepackage{colortbl}

\definecolor{wacvblue}{rgb}{0.21,0.49,0.74}
\usepackage[pagebackref,breaklinks,colorlinks,allcolors=wacvblue]{hyperref}

\title{Seeing the Unseen: Camouflaged Object Detection Beyond the Visible Spectrum}

\author{Avi Gupta\\
Indraprastha Institute of Information Technology, Delhi\\
{\tt\small avig@iiitd.ac.in}
\and
Trasha Gupta\\
Delhi Technological University, Delhi\\
{\tt\small trashagupta@dtu.ac.in}
}

\begin{document}
\maketitle
\begin{abstract}
Recent advances in camouflaged object detection (COD) have led to substantial progress in challenging low-visibility scenarios, with pioneering studies demonstrating notable success in localizing objects in camouflaged scenes. Despite these achievements, existing approaches predominantly rely on conventional three-channel RGB imagery, thereby constraining the available visual information to a limited spectral range. Multispectral images offer a wide range of information about a scene by capturing fine-grained spectral signatures. Hence, by leveraging multispectral images for COD, we introduce a novel approach to detect camouflaged objects from the corresponding multispectral inputs. In particular, we propose an end-to-end framework, \textbf{\textit{MSFormer}}, that takes a multispectral camouflaged image as input and predicts a binary mask for it. Additionally, we also provide empirical justification for integrating multispectral bands for this complex low-vision task. Our extensive experiments demonstrate the effectiveness of our method, which outperforms existing methods. Code is available \href{https://github.com/avigupta2798/MSFormer}{here}.
\end{abstract}    
\section{Introduction}
\label{sec:intro}

Camouflaged Object Detection (COD) addresses the perceptual challenge of segmenting objects that closely resemble their surroundings~\cite{10379651, Ge2024FeatureawareAI, ZHANG2023103719}. Unlike salient object detection, where objects are clearly visible, camouflaged objects rely on background matching and disruptive coloration to conceal their presence. 
This task is critical in applications requiring a robust understanding in complex environments~\cite{sun2021c2fnet, chen2022camouflaged, 10661228, DBLP:conf/cvpr/Lv0DLLBF21}, including wildlife conservation and polyp segmentation in medical imaging. However, the intrinsic ambiguity of camouflage renders standard segmentation methods ineffective.

\begin{figure}[t]
    \centering
    \includegraphics[width=\columnwidth]{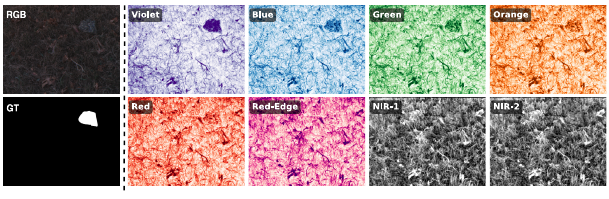}
    \caption{Visualization of a Multispectral camouflaged image. We provide a single camouflaged image in its traditional RGB form along with its corresponding ground truth. Furthermore, the corresponding multispectral image is visualized in 8 channels.}
    \label{fig:teaser}
\end{figure}

\begin{figure*}[ht]
    \centering
    \includegraphics[width=\textwidth]{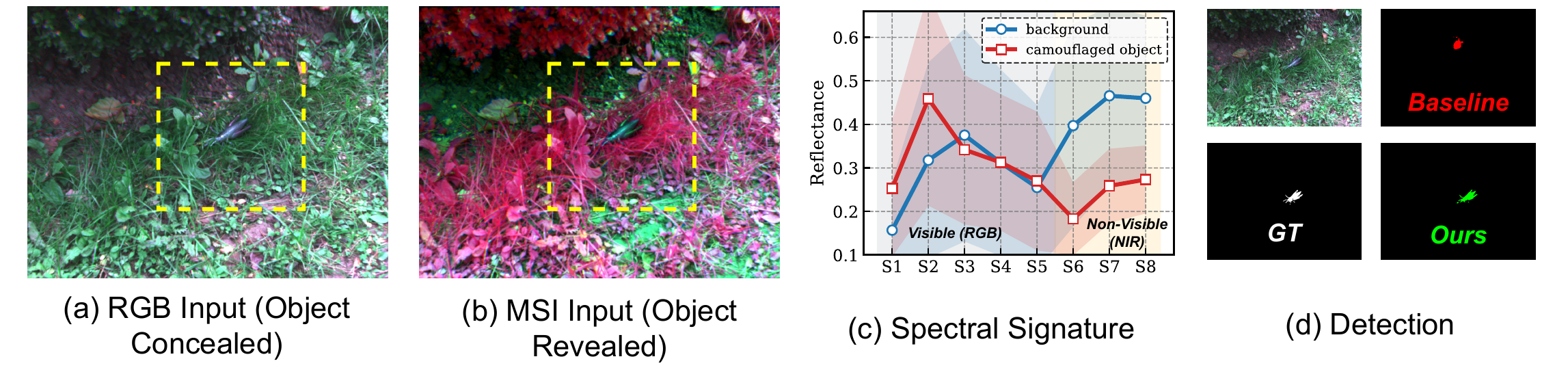}
    \caption{Camouflaged image outside the visible spectrum. (a) A camouflaged object that is nearly indistinguishable from its background in the RGB image. (b) The image becomes clearly separable in the multispectral domain. (c) The per-band reflectance is averaged over the object and background regions; the two spectral signatures overlap across the visible bands but diverge in the near-infrared bands. (d) Using this signal, our MSFormer accurately segments the object where a strong RGB baseline fails.}
    \label{fig:teaser_new}
\end{figure*}

While traditional COD methods~\cite{pang2022zoom, he2023weakly, niu2024minet, wang2024ipnet, DBLP:journals/tmm/ZhouWC24, 10379651, gupta2025circod} have achieved remarkable success, they remain constrained by the visible RGB spectrum. Multispectral imaging offers a paradigm shift by capturing distinct spectral information that remains imperceptible in the visible band. The advantage of fusing RGB with this additional spectral information has been well-validated in urban scene parsing and autonomous driving~\cite{DBLP:conf/mm/JiLBZ023, 8206396, DBLP:journals/tase/SunZYWL21, DBLP:journals/tmm/ZhouLLYH22}. Fig.\textbf{~\ref{fig:teaser}} visualizes a single MCOD sample across its eight spectral channels, providing more diverse information about the scene than a traditional RGB image. The result of this information for COD is explicit in our Fig.\textbf{~\ref{fig:teaser_new}}. An insect is effectively invisible in the RGB input (Fig.\textbf{~\ref{fig:teaser_new}{\color{wacvblue}a}}), but emerges in the false-color multispectral composite (Fig.\textbf{~\ref{fig:teaser_new}{\color{wacvblue}b}}). It is observed that the per-band reflectance (Fig.\textbf{~\ref{fig:teaser_new}{\color{wacvblue}c}}) across the visible bands ($S_1-S_6$), the mean reflectance of the object and its background are statistically indistinguishable, confirming that the camouflage holds in exactly the range humans and RGB sensors perceive; yet in the NIR bands ($S_7-S_8$), the two profiles diverge sharply. As a result, a model granted access to these band segments accurately localize the target, whereas a strong RGB baseline fails (Fig.\textbf{~\ref{fig:teaser_new}{\color{wacvblue}d}}). 

In the multispectral domain for COD, research is still in its infancy. The recently proposed MCOD benchmark~\cite{DBLP:journals/corr/abs-2509-15753} demonstrated that multispectral data significantly enhances detection capability, yet current baselines rely on adding a band-specific initial learnable layer, which is data-inefficient and prone to overfitting.
To overcome this, leveraging a transformer-based vision model is a promising direction. However, current backbones are pre-trained on massive RGB datasets and lack the capacity to process multispectral inputs. Directly applying these pre-trained models to multispectral COD is suboptimal because it discards crucial non-visible information. While ~\cite{Yadav_2025_BMVC} have attempted to tailor large vision models for challenging tasks using wavelet transforms, effectively fusing multispectral data into an RGB-centric backbone remains an open challenge.

To bridge this gap, we propose a transformer-based multispectral camouflaged object detection framework, \textit{MSFormer}. Firstly, we introduce a weight inflation strategy that adapts the standard transformer-based backbone~\cite{DBLP:journals/cvm/WangXLFSLLLS22} for multispectral inputs. This strategy projects multispectral inputs into the encoder's embedding space and enables the model to leverage the ``unseen" spectral clues of camouflaged objects while retaining the powerful, structural segmentation priors to generate multiscale features. Following this, we standardize the multiscale feature representations using feature-projection layers. Finally, we propose a cascaded decoder module that takes these uniform feature representations as input and aggregates them using concatenation and transposed convolution to predict the final mask.

We summarize our contributions as follows: 
\begin{enumerate}
    \item We propose a novel transformer-based Multispectral COD framework that takes the multispectral camouflaged image as input and predicts the corresponding binary mask.
    \item We introduce a weight inflation strategy that effectively fuses traditional RGB bands with additional band information, enabling the encoder to capture complementary cross-modal features.
    \item We propose a cascaded decoder module that aggregates the uniform multiscale features to predict the efficient mask for the camouflaged image.
    \item We conduct extensive experiments on different COD benchmarks, demonstrating that our method performs significantly well across RGB, multispectral, and hyperspectral baselines.
\end{enumerate}
\section{Related Works}
\label{sec:related}

\subsection{Camouflaged Object Detection}

Camouflaged object detection aims to segment objects that closely match their surroundings in appearance~\cite{sahane2026review}. 

The advent of deep learning has revolutionized this field, enabling the extraction of rich, multi-scale semantic features~\cite{10379651, Ge2024FeatureawareAI, ZHANG2023103719, sun2021c2fnet, chen2022camouflaged, he2025nested}. Recent state-of-the-art methods often adopt biologically inspired strategies~\cite{10661228, DBLP:conf/cvpr/Lv0DLLBF21} to mimic human visual perception. \cite{gupta2025circod} incorporated visual supervision to leverage co-saliency for object localization. More recently, the rise of foundation models has prompted works such as \cite{Yadav_2025_BMVC}, which adapt them using wavelet-based transformations to handle these challenging scenarios. Despite these advancements, the aforementioned methods predominantly rely on 3-channel RGB inputs. To address this, \cite{DBLP:journals/corr/abs-2509-15753} recently introduced a benchmark for multispectral COD, demonstrating that non-visible spectral bands can reveal targets hidden in the visible spectrum. Building on this insight, we propose a novel multispectral framework that leverages diverse spectral bands to enable robust representation learning.

\subsection{Multispectral Segmentation}

Multispectral Image segmentation (MIS) addresses the intrinsic limitations of RGB models by integrating spectral information, such as thermal or infrared data~\cite{8206396, DBLP:journals/tase/SunZYWL21, DBLP:conf/mm/JiLBZ023,DBLP:journals/tmm/ZhouLLYH22, shen2021camouflaged, wang2024towards, liu2026multispectral}. This fusion significantly enhances model robustness, particularly in adverse lighting conditions. Early works such as MFNet~\cite{8206396} used a dual-encoder architecture with concatenation for real-time fusion in autonomous tracking~\cite{DBLP:journals/corr/abs-2512-22263}. \cite{DBLP:journals/tase/SunZYWL21} employed a two-stage fusion to mitigate modality discrepancies, while~\cite{DBLP:conf/mm/JiLBZ023} introduced large-scale datasets to standardize robust segmentation. To further refine feature interactions, \cite{DBLP:journals/tmm/ZhouLLYH22} proposed attention-based mechanisms—specifically, feature-enhanced attention and multiscale fusion, respectively, to dynamically weigh the importance of different bands relative to RGB context. Recently, several studies~\cite {hupel2022adopting, hupel2025optimized, wang2025causal, zhao2025camouflage} have focussed on hyperspectral-based methods for camouflage detection in multispectral imagery.

\begin{figure}[ht]
\centering
  \begin{subfigure}[b]{0.49\columnwidth}
    \centering
    \includegraphics[width=\columnwidth]{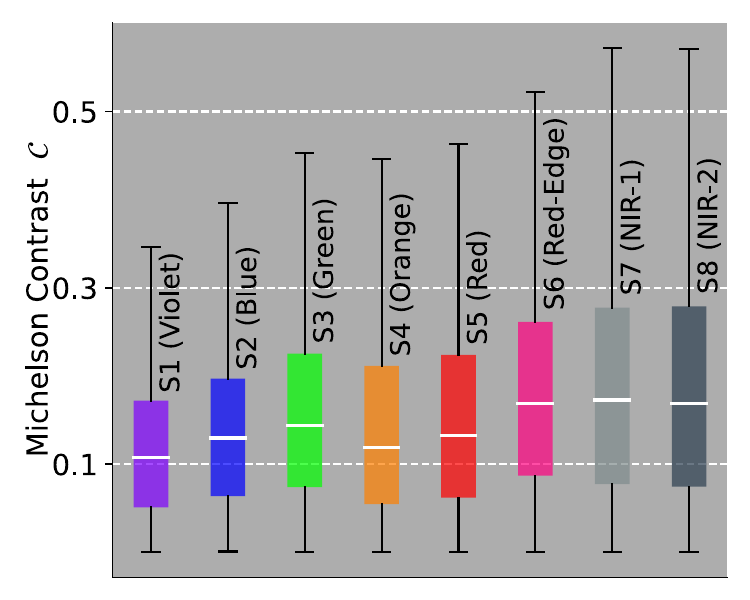} 
    \caption{} 
    \label{fig7:box} 
    \vspace{4ex}
  \end{subfigure}
  \begin{subfigure}[b]{0.49\columnwidth}
    \centering
    \includegraphics[width=\columnwidth]{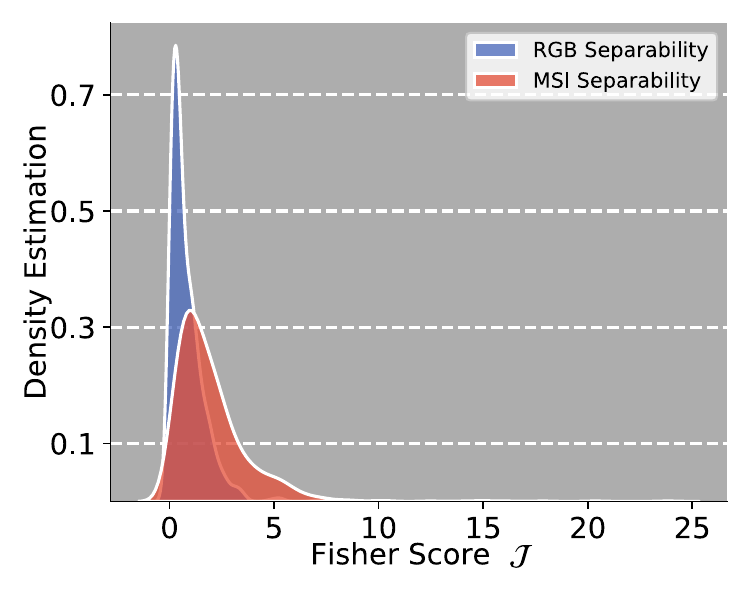} 
    \caption{} 
    \label{fig7:kde} 
    \vspace{4ex}
  \end{subfigure} 
  \begin{subfigure}[b]{0.49\columnwidth}
    \centering
    \includegraphics[width=\columnwidth]{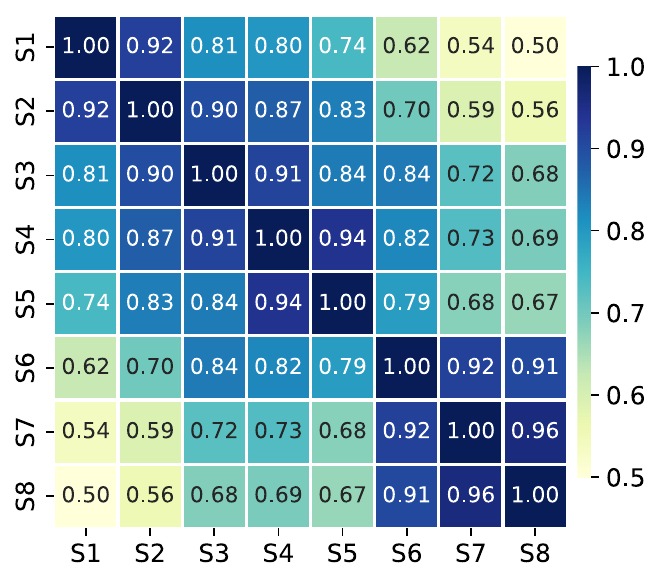} 
    \caption{} 
    \label{fig7:heatmap} 
  \end{subfigure}
  \begin{subfigure}[b]{0.49\columnwidth}
    \centering
    \includegraphics[width=\columnwidth]{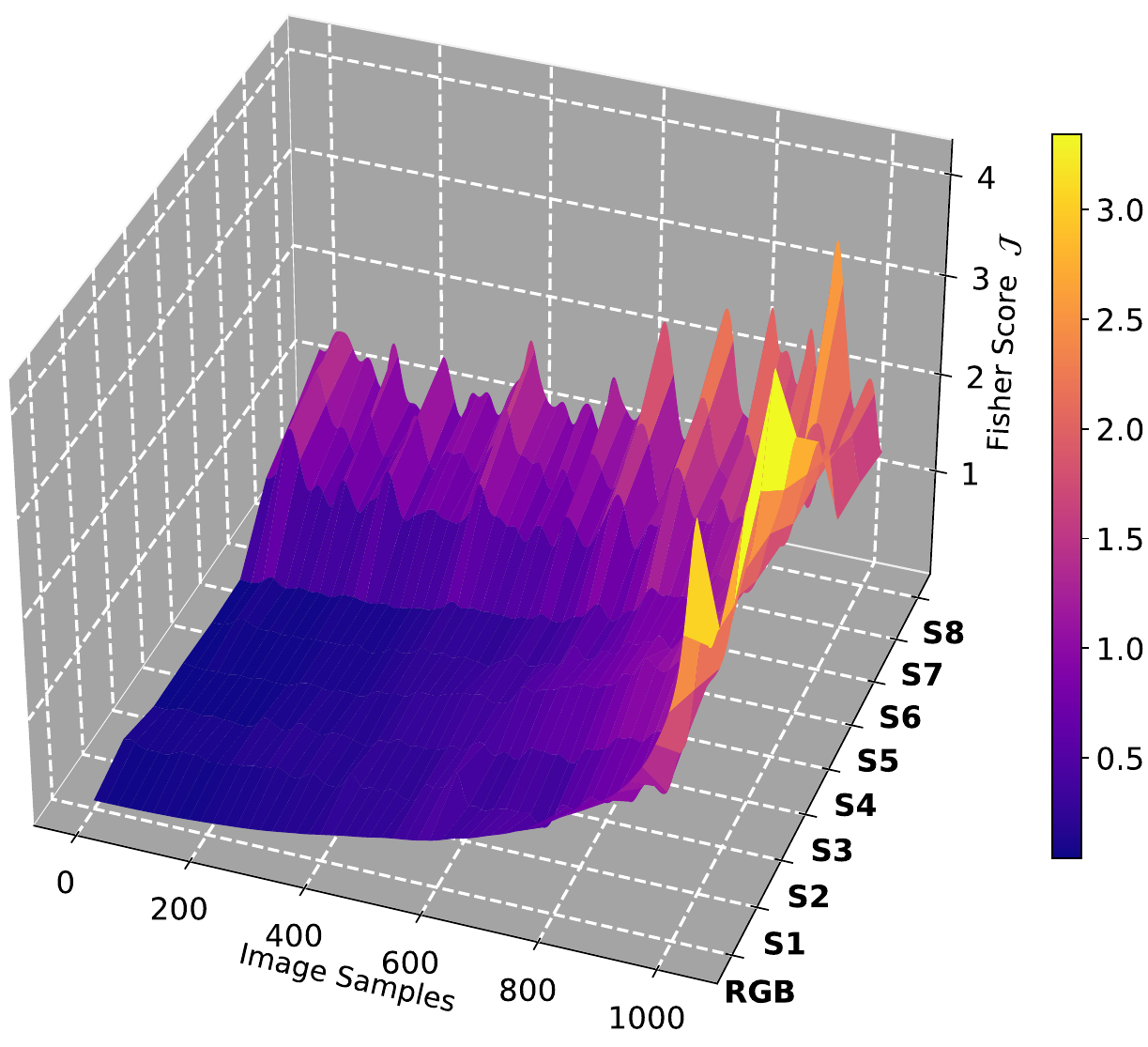} 
    \caption{} 
    \label{fig7:d} 
  \end{subfigure} 
  \caption{Statistical motivation for Multispectral COD. a) \textbf{Global Contrast Distribution: }Box plots show median contrast in NIR bands (S7, S8) is $2\times$ higher than visible bands (S5, S3, and S2). b) \textbf{Feature Separability Density: }The heavy tail in the MSI distribution (red) indicates a subset of "easy-to-segment" samples that are otherwise indistinguishable in RGB (blue). c) \textbf{Correlation Heatmap:} Low correlation between Visible and NIR bands confirms non-redundancy. d) \textbf{Spectral Landscape: }3D visualization showing consistently higher separability scores in NIR bands across the dataset.}
  \label{fig7} 
\end{figure}

\begin{figure*}[ht]
    \centering
    \includegraphics[width=\textwidth]{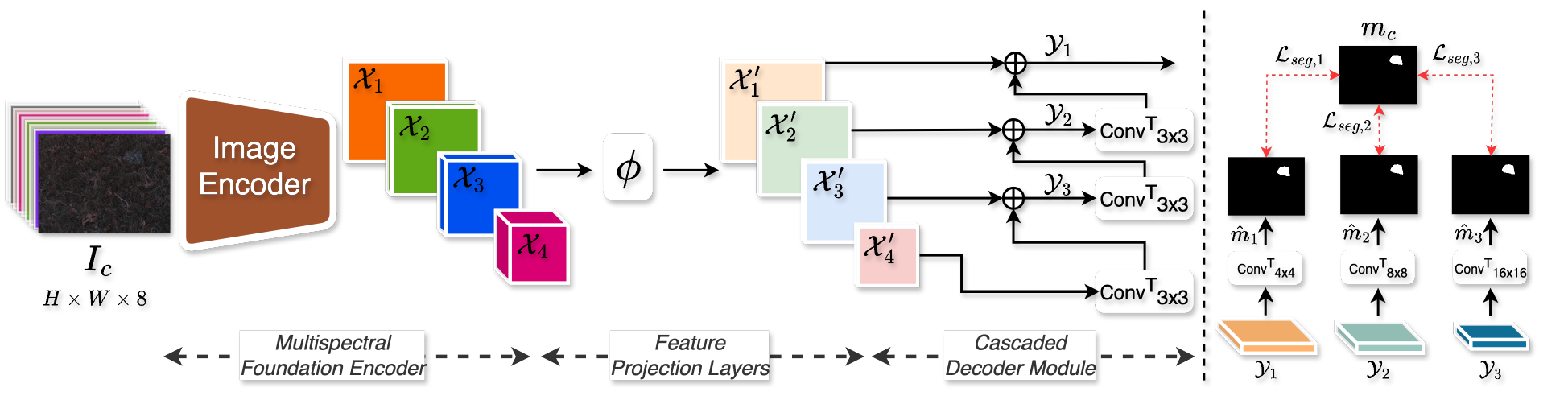}
    \caption{\textbf{Overview of our MSFormer framework.} The multispectral camouflaged image $I_c$ is passed through the image encoder that extracts the multiscale features $\mathcal{X}$. These features are standardized to a uniform channel dimension $\mathcal{X'}$ using Feature Projection Layers $\phi$, which are then fused recursively using the Cascaded Decoder Module (CDM). The hierarchical cascaded features $\mathcal{Y}$ are subsequently upsampled to restore spatial resolution and generate prediction maps at distinct scales, $\hat{m}$. Finally, we calculate the loss $\mathcal{L}_{seg}$ between predicted maps and the ground truth $m_c$.}
    \label{fig:arch}
\end{figure*}

\section{Methodology}
\label{sec:method}

\subsection{RGB \textit{vs} MSI Band}
To empirically justify the integration of multispectral imaging (MSI) for Camouflaged Object Detection (COD), we perform a rigorous statistical analysis comparing the representations of standard RGB modalities with those of the entire multispectral bands. Our analysis focuses on three key discriminative properties: \textit{optical contrast}, \textit{feature separability}, and \textit{information orthogonality}.
\subsubsection{Optical Contrast and Spectral Divergence:} The fundamental challenge in COD is the camouflaged foreground that mimics the chromaticity of the background. We quantify this by computing the Michelson Contrast ($\mathcal{C}$) for each spectral band across the entire dataset. Fig.\textbf{~\ref{fig7:box}} visualizes the global contrast distribution. We observe that in the visible RGB bands (S5, S3, and S2), the median contrast is lower, confirming the effectiveness of camouflage in the human visual range. However, a significant divergence is observed in the Near-Infrared (NIR) bands (S7–S8). The median contrast in these bands is approximately $2\times$ higher than in the visible spectrum.
\subsubsection{Feature Separability and Decision Boundaries. }Beyond raw contrast, we analyze the statistical separability of object-background distributions using the Fisher Discriminant Ratio ($\mathcal{J}$). We visualize this in two complementary forms:

\textbf{The Spectral Landscape: } Fig.\textbf{~\ref{fig7:d}} visualizes the separability score $\mathcal{J}$ as a 3D surface across the dataset samples. The topology reveals a distinct ``valley" in the RGB domain (Row 0), indicating a consistently challenging optimization landscape. Conversely, the NIR bands (Rows 7–8) form a prominent ``ridge", demonstrating that the discriminative signal in MSI is not sporadic but statistically consistent across diverse samples.

\textbf{Separability Density: } Quantitatively, Fig.\textbf{~\ref{fig7:kde}} compares the density estimation of $\mathcal{J}$. The RGB distribution (blue) is strongly peaked near zero ($\mathcal{J}$ $\approx$ 0.2), indicating that the foreground and background are statistically indistinguishable. In contrast, the MSI distribution (red) exhibits a pronounced heavy tail, extending to $\mathcal{J}>$ 5. This indicates that MSI shifts the problem from semantic inference in a low-contrast regime to feature discrimination in a high-contrast regime, effectively reducing the complexity of the underlying decision boundary.
\subsubsection{Information Orthogonality. }Here, we compute the inter-channel Pearson-correlation coefficient matrix, averaged across the dataset (Fig.~\textbf{\ref{fig7:heatmap}}). We observe high collinearity ($\rho>$ 0.82) among the visible bands (S2–S5), suggesting significant redundancy in standard RGB inputs. However, the correlation between visible and NIR bands (S7–S8) drops significantly. This validates that the multispectral bands provide orthogonal information, introducing unique variance that allows the network to resolve ambiguities present in the visible spectrum.

\subsection{Overview}
Let the input multispectral camouflaged image be denoted as $I_c \in \mathbb{R}^{H \times W \times C_{ms}}$, where $C_{ms}$ represents the number of spectral bands (comprising RGB, NIR, and other bands). $H$ and $W$ represent the image's height and width, respectively. The objective is to predict the binary segmentation map $\hat{m}_c \in [0, 1]^{H \times W}$, where each pixel indicates the probability of belonging to a camouflaged object. We propose the \textbf{M}ulti\textbf{S}pectral Trans\textbf{former} Network (\textit{MSFormer}), a hierarchical framework for extracting and fusing robust semantic features from generalized multispectral data. As illustrated in Fig.~\textbf{\ref{fig:arch}}, the architecture comprises three key components: \textit{\textbf{Multispectral Feature Extractor}}: an encoder backbone that extracts multi-scale features from $C_{ms}$-channel inputs, \textbf{\textit{Feature Projection Layers}}: standardize the channel dimensions of the extracted hierarchical features, followed by \textit{\textbf{Cascaded Decoder Module (CDM)}}: a top-down aggregation pathway that progressively recovers spatial resolution while integrating semantic context from deeper layers.

\subsection{Multispectral Feature Extractor}
To capture the long-range dependencies and robust spectral features, we employ the pyramid vision transformer (PVT)~\cite{DBLP:journals/cvm/WangXLFSLLLS22} encoder as our backbone. Unlike standard CNNs, PVT utilizes a transformer-based pyramid structure, which is highly effective for dense prediction tasks.

\paragraph{\textit{Weight Inflation Strategy:}}Standard PVT encoders are configured for three-channel RGB input. To adapt this backbone for multispectral data while preserving the robust feature representations learned during pre-training, we employ a channel-wise weight inflation strategy within the patch embedding layer. 
Let $\mathbf{W}_{rgb} \in \mathbb{R}^{K \times 3 \times P \times P}$ denote the pre-trained weights corresponding to the traditional Red ($R$), Green ($G$), and Blue ($B$) channels. We initialize the generalized weight tensor, $\mathbf{W}_{ms} \in \mathbb{R}^{K \times C_{ms} \times P \times P}$ (where $C_{ms}$ represents the total number of multispectral channels), by mapping the $\mathbf{W}_{rgb}$ weights to their corresponding R, G, and B channels. For the remaining spectral channels, we compute the arithmetic mean across the three original RGB channels. This initialization strategy is mathematically formulated as follows:
\begin{equation}
\mathbf{W}_{ms}[:, c, :, :] =
\begin{cases}
\mathbf{W}_{rgb}[:, c, :, :] & \text{if } c \in \{R, G, B\} \\
\frac{1}{3} \sum_{j=0}^{2} \mathbf{W}_{rgb}[:, j, :, :] & \text{otherwise}
\end{cases}
\end{equation}

where $c$ denotes the specific channel index, $K$ denotes the embedding dimension, and $P$ denotes the spatial dimension of the initial layer of the encoder. This initialization preserves the learned RGB representations while providing a balanced initialization for the novel spectral bands. The encoder extracts four hierarchical feature maps, denoted as $\mathcal{X}_i\in\mathbb{R}^{{\frac{H}{2^{i+1}}} \times {\frac{W}{2^{i+1}}} \times {C_i}}$, where $C_i$ is the $i^{th}$ element in $C\in\{64, 128, 320, 512\}$.

It is noteworthy that this strategy can also be adapted for n-channel bands in the hyperspectral domain.

\subsection{Cascaded Decoder Module}

The diverse scales of camouflaged objects require effective fusion of high-level semantics (for localization) and low-level details (for boundary refinement). We introduce the Cascaded Decoder Module (CDM) to aggregate the extracted features ($\mathcal{X}$) via a top-down pathway. First, the features are passed through Feature Projection Layers ($\phi$) to map them to a uniform channel dimension:
\begin{equation}
  \mathcal{X}'_i = \phi_i(\mathcal{X}_i); \quad i \in \{1, 2, 3, 4\}  
\end{equation}
where $\phi(\cdot)=$ BN(ReLU(Conv$_{3\times 3}$($\cdot$))) that resizes each extracted feature to 64 channel size ($\mathcal{X}'$). The CDM then recursively fuses these features. Starting from the deepest feature $\mathcal{X}'_4$, we employ a sequence of upsampling and concatenation operations. Let $\mathcal{Y}_i$ denote the decoded feature at stage $i$. The aggregation process is defined as:
\begin{equation}
\begin{aligned}
\mathcal{Y}_3 &= \text{Conv}^{T}_{3 \times 3}(\mathcal{X}'_4) \oplus \mathcal{X}'_3 \\
\mathcal{Y}_2 &= \text{Conv}^{T}_{3 \times 3}(\mathcal{Y}_3) \oplus \mathcal{X}'_2 \\
\mathcal{Y}_1 &= \text{Conv}^{T}_{3 \times 3}(\mathcal{Y}_2) \oplus \mathcal{X}'_1
\end{aligned}
\end{equation}

where $\text{Conv}^{T}_{3\times 3}(\cdot)$ denotes 2d-transposed convolution of $3\times 3$ kernel and $\oplus$ denotes channel-wise concatenation. This cascaded design ensures that stronger semantic cues from deeper layers guide the refinement of shallower, detail-rich features.

These hierarchical cascaded features are later upsampled to restore spatial resolution for gradient optimization using a Conv operation to get the corresponding hierarchical mask $\hat{m}$ as:
\begin{equation}
    \hat{m}_i = \text{Conv}^T_{k\times k}(\mathcal{Y}_i);\  k=2^{(i+1)}\ i\in \{1,2,3\},
\end{equation}

\subsection{Objective Function}
For our proposed approach, we employ a deep supervision strategy, evaluating camouflaged prediction maps at distinct scales: $\hat{m}_1$ (finest), $\hat{m}_2$, and $\hat{m}_3$ (coarsest). Each prediction map consists of a single-channel probability map $\hat{m} \in \mathbb{R}^{H \times W}$. The total training objective $\mathcal{L}_{total}$ is the sum of the losses at each scale:
\begin{equation}
\mathcal{L}_{total} = \sum_{k=1}^{3} \mathcal{L}_{seg,k}(\hat{m}_k, m_c)
\end{equation}

where $m_c$ is the ground truth binary mask. The segmentation loss $\mathcal{L}_{seg}$ is a combination of the Weighted Binary Cross-Entropy (BCE) loss and the Weighted Intersection-over-Union (IoU) loss to handle the class imbalance inherent in camouflaged scenes:
\begin{equation}
\mathcal{L}_{seg}(\hat{m}, m_c) = \mathcal{L}_{BCE}^w(\hat{m}, m_c) + \mathcal{L}_{IoU}^w(\hat{m}, m_c)
\end{equation}

where $\hat{m}\in\{\hat{m}_1, \hat{m}_2, \hat{m}_3\}$. By combining these two losses, we can effectively supervise both the pixel-level details and the foreground-background regions.
This multi-scale supervision forces the network to learn global object localization at deeper layers while refining fine-grained boundaries at shallower layers.
\section{Experiments}
\label{sec:experiments}

\subsection{Experimental Details}
\subsubsection{Datasets:}We consider the current MCOD benchmark dataset \cite{DBLP:journals/corr/abs-2509-15753} to evaluate our proposed approach. Unlike the traditional three-channel RGB band, MCOD consists of multispectral bands. The dataset comprises 1,527 multispectral images, divided into 1,027 training and 500 test images, with each image containing eight channels. Additionally, we evaluate on the recently proposed hyperspectral dataset for COD (HyperCOD)~\cite{DBLP:conf/aaai/BaiXLQBCPL26}. The HyperCOD dataset comprises 350 hyperspectral images with 200 distinct spectral bands. The dataset is divided into 280 training samples and 70 testing samples. To further analyze the robustness of our proposed framework, we evaluate MSFormer on traditional COD benchmarks, 4,040 (COD10K~\cite{Fan_2020_CVPR} + CAMO~\cite{ltnghia-CVIU2019}) training images, 2,026 COD10K testing images, 250 CAMO testing images, and 4,121 NC4K~\cite{DBLP:conf/cvpr/Lv0DLLBF21} testing images.  
\subsubsection{Evaluation Metrics:}Following~\cite{DBLP:journals/corr/abs-2509-15753}, we evaluate our approach on standard metrics: Mean Absolute Error ($\mathcal{M}$), mean F-measure ($\mathcal{F}_{\beta}$)~\cite{DBLP:conf/cvpr/AchantaHES09}, adaptive F-measure ($\alpha\mathcal{F}$)~\cite{DBLP:conf/cvpr/AchantaHES09}, S-measure($\mathcal{S}_m$)~\cite{DBLP:conf/iccv/FanCLLB17}, and Enhanced Alignment Measure ($\mathcal{E}_\xi$)~\cite{DBLP:conf/ijcai/FanGCRCB18}.
\subsubsection{Implementation Details:}We implemented our framework
using the PyTorch framework on a single workstation of
NVIDIA 4090 GPU. We adopt a pre-trained PVT-V2~\cite{DBLP:journals/cvm/WangXLFSLLLS22} model as the backbone. The network parameters are optimized using the AdamW optimizer, with an initial learning rate of 5e-5, weight decay of 0.0001, a batch size of 8, and 65 epochs. All training and testing images are resized to $512 \times 512$ and augmented with random flips, mirroring, and rotations. We will release the code for reproducibility.

\begin{table}[h]
\centering
\caption{Comparison of methods on MCOD benchmark dataset. $\uparrow/\downarrow$ denotes the larger/smaller is better. Best results are marked in \textbf{Bold}. The results are excerpted from~\cite{DBLP:journals/corr/abs-2509-15753}}
\renewcommand{\arraystretch}{1} 
\resizebox{\columnwidth}{!}{
\setlength{\tabcolsep}{10pt}
\begin{tabular}{l c c c c}
\toprule
Models & $\mathcal{E}_{\xi} \uparrow$ & $\mathcal{S}_{m} \uparrow$     & $\mathcal{F}_{\beta} \uparrow$  & $\mathcal{M} \downarrow$ \\ \midrule
SINet~\cite{Fan_2020_CVPR} & 0.758 & 0.616 & 0.369 & 0.006 \\
LSR~\cite{DBLP:conf/cvpr/Lv0DLLBF21} & 0.830 & 0.625  & 0.373  & 0.005 \\
CODCEF~\cite{DBLP:journals/sensors/HuangLZW21} & 0.763  & 0.677  & 0.444  & 0.004 \\
C2FNet~\cite{sun2021c2fnet} & 0.726  & 0.721  & 0.403  & 0.010 \\
C2FNet-V2~\cite{chen2022camouflaged} & 0.913  & 0.810  & 0.654  & 0.008 \\
SINet-V2~\cite{9444794} & 0.849  & 0.728  & 0.492  & 0.004 \\
ASBI~\cite{ZHANG2023103719} & 0.684  & 0.675  & 0.370  & 0.014 \\
FIRNet~\cite{Ge2024FeatureawareAI} & 0.882  & 0.738  & 0.537  & 0.004 \\
PRNet~\cite{10379651} &  \underline{0.926}  & 0.826  &  \underline{0.698}  & \textbf{0.002} \\
IdeNet~\cite{10661228} & 0.846  & 0.808  & 0.588  & 0.004 \\
PCNet~\cite{plantcamo} & 0.633  & \underline{0.855}  & 0.386  &  \underline{0.003} \\ \midrule
\rowcolor{wacvblue!10}
Ours &  \textbf{0.972} &  \textbf{0.881} &  \textbf{0.805} &  \textbf{0.002}  \\ \bottomrule
\end{tabular}
}
\label{tab: main}
\end{table}
\begin{table}[h]
\centering
\caption{Comparison of results under RGB and MSI inputs on MCOD benchmark. $\uparrow/\downarrow$ denotes the larger/smaller is better. Best results are marked in \textbf{Bold}. The results are excerpted from~\cite{DBLP:journals/corr/abs-2509-15753}}
\renewcommand{\arraystretch}{1} 
\resizebox{\columnwidth}{!}{
\setlength{\tabcolsep}{7pt}
\begin{tabular}{lc|cccc}
\toprule
Models & Inputs & $\mathcal{E}_{\xi} \uparrow$ & $\mathcal{S}_{m} \uparrow$     & $\mathcal{F}_{\beta} \uparrow$  & $\mathcal{M} \downarrow$ \\ \midrule

\multirow{2}{*}{SINet~\cite{Fan_2020_CVPR}} & RGB & 0.695  & 0.601  & 0.335  & 0.005 \\
 & MSI & 0.758 & 0.616 & 0.369 & 0.006 \\ \midrule
\multirow{2}{*}{CODCEF~\cite{DBLP:journals/sensors/HuangLZW21}} & RGB & 0.718  &  0.632   & 0.359   & 0.005 \\
 & MSI & 0.763  & 0.677  & 0.444  & 0.004 \\ \midrule
\multirow{2}{*}{C2FNet-V2~\cite{chen2022camouflaged}} & RGB & 0.885   & 0.743   & 0.553   & 0.009 \\
 & MSI & 0.913  & 0.810  & 0.654  & 0.008 \\ \midrule
\multirow{2}{*}{PCNet~\cite{plantcamo}} & RGB & 0.397   & 0.788  & 0.149   & 0.005 \\
 & MSI & 0.633  & 0.855  & 0.386  & 0.003 \\ \midrule

\rowcolor{wacvblue!10}
 & RGB &  \underline{0.950} &  \underline{0.825} &  \underline{0.707} &  \underline{0.003}  \\
\rowcolor{wacvblue!10}
\multirow{-2}{*}{Ours} &  MSI &  \textbf{0.972} & \textbf{0.881} & \textbf{0.805} &  \textbf{0.002}  \\ \bottomrule
\end{tabular}
}
\label{tab: rgbvsmsi}
\end{table}
\subsection{Experimental Results \& Further Analysis}
\subsubsection{Comparison with State-of-the-Art}
Following~\cite{DBLP:journals/corr/abs-2509-15753}, we compare our proposed MSFormer with eleven COD models on the MCOD dataset. As illustrated in Table~\ref{tab: main}, MSFormer outperforms all the previous works by a significant margin. Our proposed approach bridges the performance gap by effectively leveraging multi-spectral cues, achieving state-of-the-art results across all metrics. Particularly, MSFormer achieves the $\mathcal{F}_\beta$ of 0.805, outperforming \cite{10379651} by $\approx$15\%. Similarly across other metrics, $\mathcal{E}_{\xi}$, $\mathcal{S}_m$, and $\mathcal{M}$, our approach surpasses current state-of-the-art~\cite{10379651, plantcamo} by $\approx$2\%. These results demonstrate that incorporating multispectral information yields better scene understanding, with more accurate object localization and finer details.

To further assess the performance of our proposed approach, we compare it with that of multispectral inputs versus traditional RGB inputs. The results, summarized in Table~\ref{tab: rgbvsmsi}, demonstrate that MSFormer not only surpasses prior work but also that the multispectral input shows consistent improvements over the RGB input. This analysis demonstrates the effectiveness of multispectral data in precisely detecting camouflaged objects.

\begin{figure*}[ht]
    \centering
    \includegraphics[width=\textwidth]{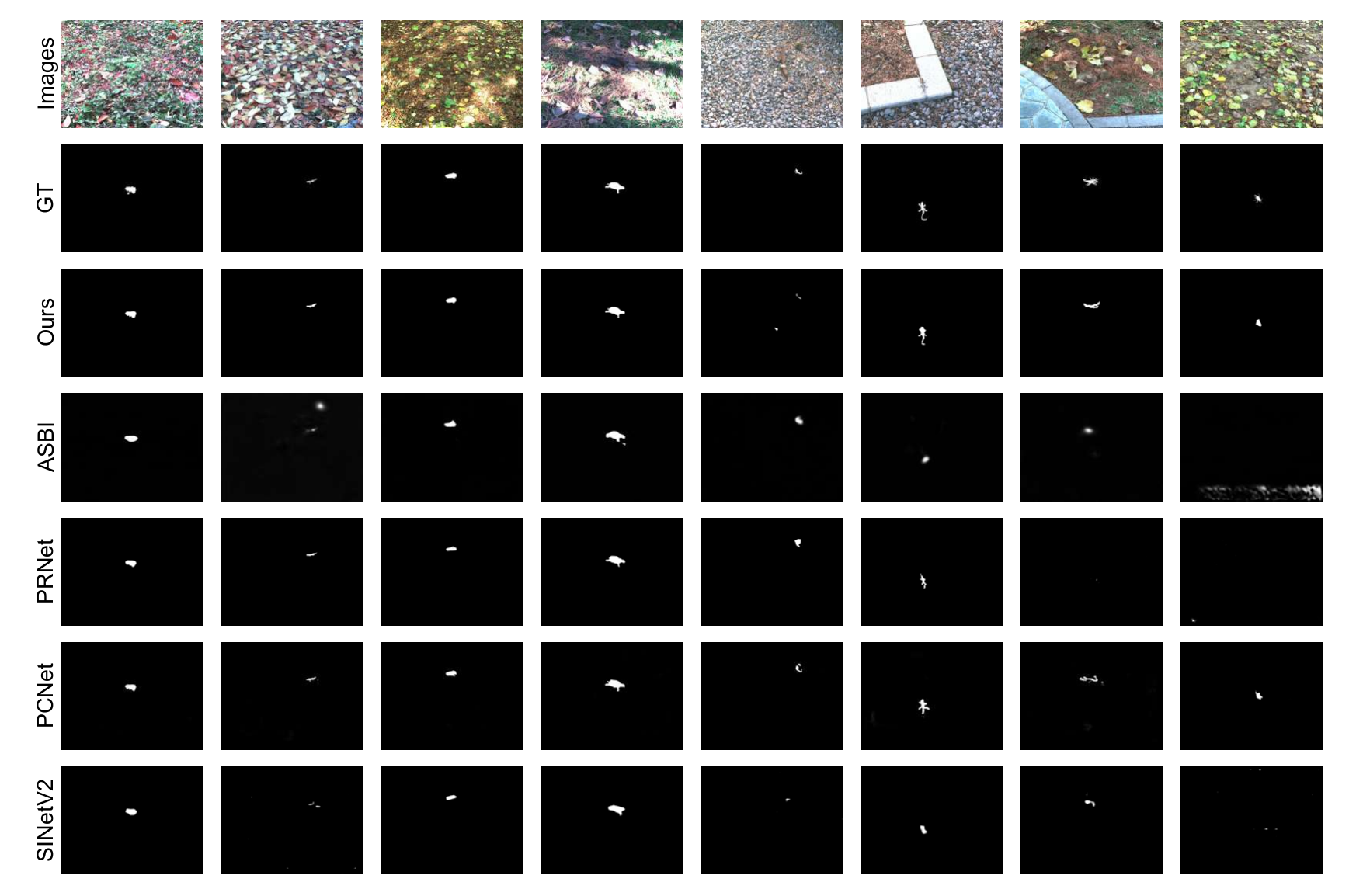}
    \caption{Qualitative comparison of our proposed approach with other COD methods.}
    \label{fig:vis}
\end{figure*}

The qualitative results of MSFormer, compared with other COD methods, are visualized in Fig.\textbf{~\ref{fig:vis}}. As shown in the figure, MSFormer produces accurate segmentation masks in challenging scenarios. Compared to other methods, MSFormer provides segmentation results with more
accurate object boundaries.

\begin{table*}[ht]
\centering
\caption{Comparison of methods on the traditional COD benchmark dataset. $\uparrow/\downarrow$ denotes the larger/smaller is better. Best results are marked in \textbf{Bold}.}
\setlength{\tabcolsep}{7pt}
\renewcommand{\arraystretch}{1} 
\resizebox{\textwidth}{!}{
\begin{tabular}{l cccc cccc cccc}
\toprule \midrule
\multirow{2}{*}{Methods} & \multicolumn{4}{c}{CAMO} & \multicolumn{4}{c}{COD10K} & \multicolumn{4}{c}{NC4K} \\ 
\cmidrule(lr){2-5} \cmidrule(lr){6-9} \cmidrule(lr){10-13}
 & $\mathcal{S}_{m} \uparrow$ & $\mathcal{F}_{\beta} \uparrow$  & $\mathcal{M} \downarrow$  & $\mathcal{E}_{\xi} \uparrow$ 
 & $\mathcal{S}_{m} \uparrow$ & $\mathcal{F}_{\beta} \uparrow$  & $\mathcal{M} \downarrow$  & $\mathcal{E}_{\xi} \uparrow$
 & $\mathcal{S}_{m} \uparrow$ & $\mathcal{F}_{\beta} \uparrow$  & $\mathcal{M} \downarrow$  & $\mathcal{E}_{\xi} \uparrow$ 
 \\ \midrule

SINet~\cite{Fan_2020_CVPR} & 0.745  &0.644 & 0.092  & 0.804 & 0.776  &0.631 & 0.043  & 0.864 & 0.808  &0.723 & 0.058  & 0.871 \\
ZoomNet~\cite{pang2022zoom} & 0.820  &0.752 & 0.066  & 0.892 & 0.838  &0.729 & 0.029  & 0.911 & 0.853  &0.784 & 0.043  & 0.896 \\
CRNet~\cite{he2023weakly} & 0.735  & 0.641 & 0.092  & 0.815 & 0.733  &0.576 & 0.049  & 0.832 & 0.775  &0.688 & 0.063  & 0.855 \\
MiNet~\cite{niu2024minet} & 0.813  &0.751 & 0.068  & 0.881 & 0.794  &0.683 & 0.036  & 0.889 & 0.828  &0.749 & 0.048  & 0.900 \\
IPNet~\cite{wang2024ipnet} & 0.841  &0.793 & 0.051  & 0.918 & 0.825  &0.709 & 0.029  & 0.910 & 0.860  &0.798 & 0.039  & 0.922 \\
DINet~\cite{DBLP:journals/tmm/ZhouWC24} & 0.821  & 0.790 & 0.068  & 0.874 & 0.832  & 0.744 & 0.031  & 0.903 & 0.856  & 0.820 & 0.043  & 0.909 \\
PRNet~\cite{10379651} & 0.872  & 0.831 & 0.050  & 0.922 & 0.874  & 0.799 & \underline{0.022}  & 0.937 & 0.891  & 0.848 & 0.031  & 0.935 \\
MCSWA-Net~\cite{DBLP:journals/tmm/SongYLHL26} & \textbf{0.877}  & \underline{0.841} & \underline{0.046}  & \underline{0.926} & \textbf{0.886}  & \underline{0.817} & \textbf{0.020}  & \underline{0.940} & \underline{0.894}  & \underline{0.854} & \underline{0.031}  & \underline{0.937} \\
\midrule
\rowcolor{wacvblue!10}
Ours & \underline{0.875}  & \textbf{0.866} & \textbf{0.045}  & \textbf{0.929} & \underline{0.876}  & \textbf{0.835} & \textbf{0.020}  & \textbf{0.948} & \textbf{0.898}  & \textbf{0.877} & \textbf{0.030}  & \textbf{0.943}  \\ 
\midrule
\bottomrule
\end{tabular}
}
\label{tab: results_cod}
\end{table*}

\subsubsection{Analysis on Traditional COD Benchmarks.}
In this section, we evaluate our proposed approach on three standard RGB camouflaged object detection benchmarks, CAMO, COD10K, and NC4K. Since these datasets provide only three-channel RGB input, the multispectral weight-inflation strategy is inactive and reduces to standard pre-trained initialization;
The results are reported in Table\textbf{~\ref{tab: results_cod}}. 
On both CAMO and COD10K, MSFormer attains the best score on three of the four metrics and ranks a close second on the structure measure $\mathcal{S}_{m}$. 
On CAMO, MSFormer reaches $\mathcal{F}_{\beta}$=0.866, exceeding MCSWA-Net (0.841) by 0.025 and PRNet (0.831) by 0.035 (+4.2\% relative); on COD10K, it achieves $\mathcal{F}_{\beta}$=0.835, improving on MCSWA-Net (0.817) by 0.018 and PRNet (0.799) by 0.036. 
On $\mathcal{S}_{m}$, MCSWA-Net edges ahead on both datasets. It is important to note that these differences are marginal.

\begin{table}[htbp]
\centering
\caption{Comparison of methods on HyperCOD benchmark dataset. $\uparrow/\downarrow$ denotes the larger/smaller is better. Best results are marked in \textbf{Bold}. The results are excerpted from~\cite{DBLP:conf/aaai/BaiXLQBCPL26}}
\renewcommand{\arraystretch}{1} 
\resizebox{\columnwidth}{!}{
\setlength{\tabcolsep}{10pt}
  \begin{tabular}{l c c c c c}
    \toprule
    \midrule
    Method & Setting & $\mathcal{M}$ $\downarrow$ & $\mathcal{E}_{\xi}$ $\uparrow$ & $\mathcal{S}_m$ $\uparrow$ & $\alpha\mathcal{F}$ $\uparrow$\\
    \midrule
    SINet-V2~\cite{9444794} & RGB   & 0.0033 & 0.732 & 0.746 & 0.480   \\
    ZoomNet~\cite{pang2022zoom}  & RGB  & 0.0044 & 0.831 & 0.757 & 0.352  \\
    FRINet~\cite{xie2023frequency}    & RGB & 0.0027 & \underline{0.889} & 0.759 & 0.604  \\
    HGINet~\cite{yao2024hierarchical}    & RGB & 0.0039 & 0.850 & 0.766 & 0.585 \\
    Camoformer~\cite{yin2024camoformer}  & RGB & 0.0070 & 0.673 & 0.355 & 0.660  \\
    \midrule
    SAD~\cite{zheng2023boundary}      & H-SOD & 0.1505 & 0.325 & 0.483 & 0.0061   \\
    DMSSN~\cite{qin2024dmssn}    & H-SOD  & 0.0295 & 0.687 & 0.446 & 0.409   \\
    SMN-PVT~\cite{fu2026hypervision}  & H-SOD  & 0.0066 & 0.654 & 0.531 & 0.103  \\
    Hyper-HRNet~\cite{qiu2025hsod} & H-SOD & 0.0149 & 0.818 & 0.608 & 0.150   \\
    \midrule
    HSC-SAM~\cite{DBLP:conf/aaai/BaiXLQBCPL26} & H-COD & \textbf{0.0017} & 0.853 & \textbf{0.802} &\underline{ 0.681}  \\
    \rowcolor{wacvblue!10}
    Ours & H-COD & \underline{0.0020} & \textbf{0.891} & \underline{0.768} & \textbf{0.692}  \\ \midrule
    \bottomrule
  \end{tabular}
  }
  \label{tab:hypercod_main}
\end{table}

\subsubsection{Robustness on Hyperspectral Data.}
To evaluate whether MSFormer generalizes beyond the eight-band MCOD setting, we test it on the recently proposed HyperCOD benchmark~\cite{DBLP:conf/aaai/BaiXLQBCPL26}. This setting is considerably more challenging than MCOD. 
We compare against three families of methods, summarized in Table\textbf{~\ref{tab:hypercod_main}}: conventional RGB-only COD models, hyperspectral salient-object-detection (H-SOD) methods, and hyperspectral camouflaged-object-detection (H-COD) methods.

Among the task-appropriate H-COD methods, MSFormer is highly competitive compared with state-of-the-art HSC-SAM, and the two approaches are effectively complementary across the metric set. Our method attains the best E-measure ($\mathcal{E}_{\xi}$=0.891, improving on HSC-SAM's 0.853 by 4.5\% and edging the strongest RGB model, FRINet, at 0.889) and the best adaptive F-measure ($\alpha\mathcal{F}$=0.692, surpassing HSC-SAM's 0.681 and the best RGB result, Camoformer's 0.660, by 4.8\%). 
On the remaining two metrics, HSC-SAM achieves a marginally lower MAE (0.0017 \textit{vs} 0.0020) and a higher structure measure ($\mathcal{S}_{m}$=0.802 \textit{vs} 0.768). 
This demonstrates that our adaptation strategy scales from the multispectral to the hyperspectral setting, showing that MSFormer is a general framework for spectrally rich camouflage-object detection.

\subsubsection{Effectiveness of Individual Spectral Bands.}

\begin{figure}
    \centering
    \includegraphics[width=\columnwidth]{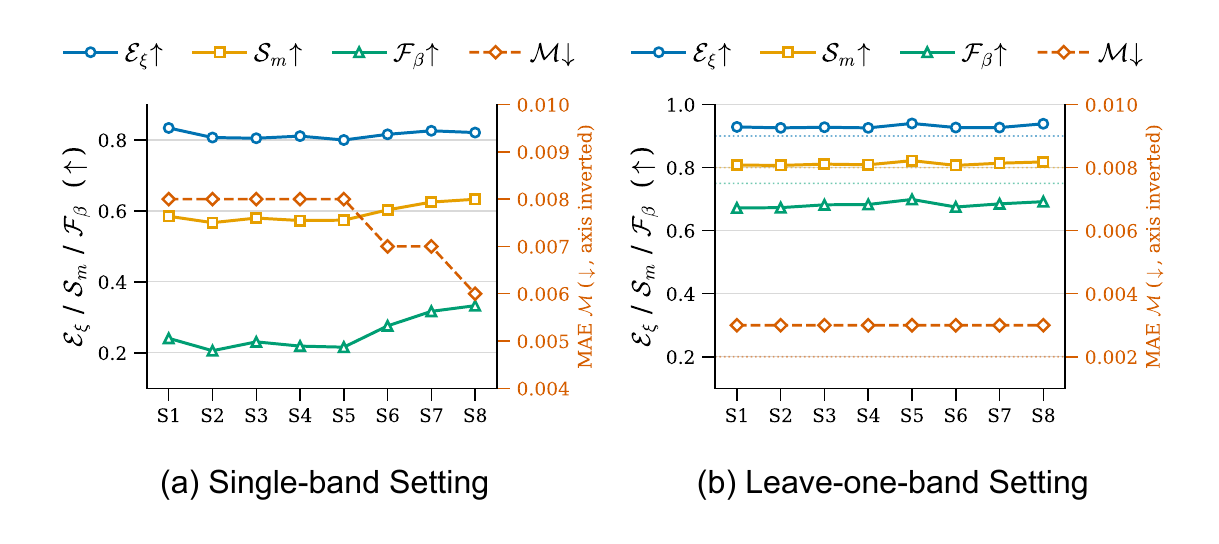}
    \caption{Per-band analysis on the MCOD benchmark. (a) Single-band input: the full pipeline is trained on one band at a time. (b) Leave-one-out: the pipeline is trained on the seven remaining bands. $\mathcal{M}$ is plotted on the right axis.}
    \label{fig:bands_ablation}
\end{figure}

To understand how the spectral bands contribute individually and jointly, we conduct two complementary experiments in which the complete framework is retrained under controlled band settings, reported in Fig.\textbf{~\ref{fig:bands_ablation}}. In the single-band setting (Fig.\textbf{~\ref{fig:bands_ablation}a}), the network is trained on a single band at a time, thereby isolating the discriminability of each channel. In the leave-one-band-out setting (Fig.\textbf{~\ref{fig:bands_ablation}b}), the network is trained on the remaining seven bands with a single band removed, exposing the marginal contribution and redundancy of that channel.

The single-band results in Fig.\textbf{~\ref{fig:bands_ablation}a} show a severe degradation across every channel relative to the full model (summarized in Table~\ref{tab: main}). 
The NIR bands ($S_7-S_8$) attain the highest $\mathcal{F}_{\beta}$, rising monotonically toward $S_8$, whereas the highly correlated visible bands ($S_1-S_6$) remain flat at the bottom. 
The leave-one-out results in Fig.\textbf{~\ref{fig:bands_ablation}b} demonstrate that removing any single band from the eight-channel input leaves performance high and remarkably flat across the choice of removed band, with no band whose removal causes a disproportionate collapse. 
Crucially, every leave-one-out configuration still falls measurably short of the full eight-band model. This persistent gap confirms that, although no band is individually critical, each band contributes a non-trivial complementary signal, and the complete multispectral stack remains strictly optimal. 

\begin{table}[h]
\centering
\caption{Ablation study for the effectiveness of architectural components on the MCOD benchmark dataset. $\uparrow/\downarrow$ denotes the larger/smaller is better. Best results are marked in \textbf{bold}.}
\renewcommand{\arraystretch}{1} 
\resizebox{\columnwidth}{!}{
\setlength{\tabcolsep}{3pt}
\begin{tabular}{cc|cccc}
\toprule
Cascaded Decoder & Feature Projection & \multirow{2}{*}{$\mathcal{E}_{\xi} \uparrow$} & \multirow{2}{*}{$\mathcal{S}_{m} \uparrow$}     & \multirow{2}{*}{$\mathcal{F}_{\beta} \uparrow$} & \multirow{2}{*}{$\mathcal{M} \downarrow$} \\
Module (CDM) & Layers & & & & \\ \midrule
\cross & \cross  & 0.707 & 0.621 & 0.303 & 0.008  \\
\cross& \checkmark & 0.812 & 0.697 & 0.482 & 0.005 \\
\checkmark & \cross & 0.953 & 0.848 & 0.742 & 0.003 \\ 
\rowcolor{wacvblue!10}
\checkmark & \checkmark &  \textbf{0.972} & \textbf{0.881} & \textbf{0.805} & \textbf{0.002} \\ \bottomrule
\end{tabular}
}
\label{tab: ablation_indiv}
\end{table}

\subsubsection{Effectiveness of individual architecture modules.}
We analyze the two core architectural components of MSFormer, the Feature Projection Layers and the Cascaded Decoder Module (CDM), to isolate their individual and joint contributions, with the results summarized in Table\textbf{~\ref{tab: ablation_indiv}}. Removing both components reduces the decoder to a direct upsampling of the deepest backbone feature, with no channel standardization and no multiscale fusion; resulting in $\mathcal{E}_{\xi}$=0.707, $\mathcal{S}_{m}$=0.621, $\mathcal{F}_{\beta}$=0.303, and $\mathcal{M}$=0.008, confirming that naively decoding the final-stage feature is fundamentally insufficient for camouflaged object detection, where fine-grained boundary recovery reflected in the sharp $\mathcal{F}_{\beta}$ collapse is especially critical.
Introducing the projection layers alone, which standardizes the four hierarchical feature maps to a uniform 64-channel representation but does not yet recursively fuse them, yields a modest but consistent improvement ($\mathcal{E}_{\xi}$ +14.9\%, $\mathcal{S}_{m}$ +12.2\%, $\mathcal{F}_{\beta}$ +59.1\%, $\mathcal{M}$ -37.5\% relative to the baseline). 
In contrast, enabling CDM alone, without the layers, yields substantially larger gains across all metrics ($\mathcal{E}_{\xi}$=0.953, $\mathcal{S}_{m}$=0.848, $\mathcal{F}_{\beta}$=0.742, $\mathcal{M}$=0.003). 
Combining both components achieves the best results across all four metrics ($\mathcal{E}_{\xi}$=0.972, $\mathcal{S}_{m}$=0.881, $\mathcal{F}_{\beta}$=0.805, $\mathcal{M}$=0.002). Notably, adding the projection layers on top of an already-enabled CDM yields its largest marginal improvement on $\mathcal{F}_{\beta}$ (+0.063, 8.5\% relative) and $\mathcal{M}$ (-0.001, 33.3\% relative reduction), while producing comparatively smaller gains on $\mathcal{E}_{\xi}$ (+2.0\%) and $\mathcal{S}_{m}$ (+3.9\%). 

\begin{table}[h]
\centering
\caption{Ablation study for the effectiveness of initialization strategy on the MCOD benchmark dataset. $\uparrow/\downarrow$ denotes the larger/smaller is better. Best results are marked in \textbf{bold}.}
\renewcommand{\arraystretch}{1} 
\resizebox{\columnwidth}{!}{
\setlength{\tabcolsep}{7pt}
\begin{tabular}{c|cccc}
\toprule
Initialization Strategy & $\mathcal{E}_{\xi} \uparrow$ & $\mathcal{S}_{m} \uparrow$     & $\mathcal{F}_{\beta} \uparrow$  & $\mathcal{M} \downarrow$ \\ \midrule
Random & 0.931 & 0.817 & 0.695 & 0.003 \\
Blue-channel& 0.930 & 0.814 & 0.689 & 0.003 \\
Red-channel& 0.935 & 0.816 & 0.689 & 0.003 \\
Green-channel& 0.938 & 0.821 & 0.696 & 0.003 \\
Zero-value & 0.953 & 0.860 & 0.769 & 0.002 \\
Average & 0.965 & 0.870 & 0.783 & 0.002 \\
\rowcolor{wacvblue!10}
Ours & \textbf{0.972} & \textbf{0.881} & \textbf{0.805} & \textbf{0.002} \\ \bottomrule
\end{tabular}
}
\label{tab: ablation_initialization}
\end{table}

\begin{figure*}[ht]
    \centering
    \includegraphics[width=\textwidth]{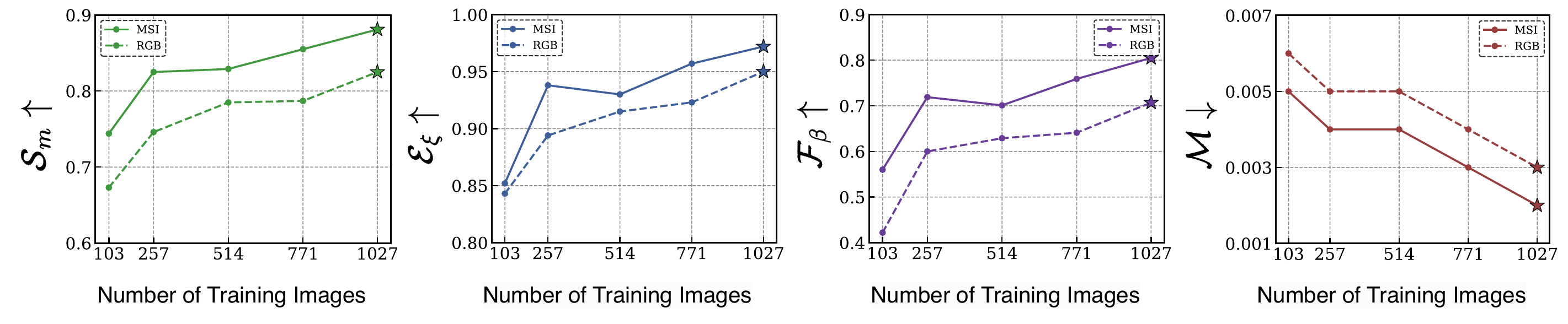}
    \caption{Data-generalization on the MCOD benchmark (with training images 103/257/514/771/1,027) using full multispectral (MSI, solid) \textit{vs} visible-only (RGB, dashed) input. Stars mark the full-data models.}
    \label{fig:data_gen}
\end{figure*}

\subsubsection{Ablation on data generalization.}
To assess how effectively MSFormer generalizes under varying amounts of supervision, we retrain the model on progressively larger subsets of the MCOD training set: 103, 257, 514, 771, and the full 1,027 images ($\approx$10\%, 25\%, 50\%, 75\%, and 100\%), under two input settings: the multispectral input (MSI) and the visible-only RGB bands. Fig.\textbf{~\ref{fig:data_gen}} reports all four metrics, where the star denotes the full data.
Across all data settings and metrics, the MSI variant dominates its RGB counterpart. With only 103 training images, MSI raises $\mathcal{F}_{\beta}$ from 0.420 to 0.557 and $\mathcal{S}_{m}$ from 0.672 to 0.745, whereas with the full 1,027 images, these gaps narrow to 13.9\% and 6.9\%, respectively. 
Notably, MSFormer trained on 25\% of the multispectral images already attains $\mathcal{F}_{\beta}$=0.722 and $\mathcal{S}_{m}$=0.823, respectively, surpassing and matching the RGB model trained on the entire dataset. 
The same trend holds for the E-measure, where MSI at 771 images ($\mathcal{E}_{\xi}$=0.958) already exceeds full-data RGB (0.950).
Importantly, MSI retains a +13.9\% $\mathcal{F}_{\beta}$ and +6.9\% $\mathcal{S}_{m}$, confirming that the benefit is representational rather than a transient low-data artifact. 
The RGB and MSI settings also differ in saturation behaviour. The RGB curves rise steeply at first, $\mathcal{F}_{\beta}$ climbs from 0.420 to 0.600 between 103 and 257 images, yet plateau well below the MSI curves, whereas the MSI curves begin near their ceiling and improve more gradually, reflecting that each multispectral image is information-rich and the model extracts more usable signal per sample. 

\subsubsection{Analysis on Different Initialization Strategies.}
In this section, we analyze different initialization strategies and identify their impact on the framework. Adapting a three-channel pretrained patch-embedding layer to the eight-channel multispectral input requires a principled scheme for initializing the five non-visible-band filters. Table\textbf{~\ref{tab: ablation_initialization}} isolates the impact of this choice while keeping the rest of MSFormer fixed. Random initialization of the bands discards any low-level structural prior for the additional channels, forcing the network to learn edge and texture filters for these bands from scratch and yielding the weakest foreground quality in the table. Furthermore, single-channel strategies that replicate the learned Blue, Red, or Green filter across all extra bands provide no benefit over random initialization. 
Zero-value initialization outperforms both random and single-channel schemes by leaving the new-band filters inactive at the start, thereby avoiding harmful bias while keeping the pretrained RGB filters fully intact. 
The Average prior improves further, because the mean of the three RGB filters supplies the new bands with a balanced, achromatic, but structurally meaningful starting point  
that transfers better than an inactive (zero) or biased (single-channel) filter. Our proposed inflation strategy achieves the best results on every metric, improving over random initialization by 4.4\% in $\mathcal{E}_{\xi}$, 7.8\% in $\mathcal{S}_{m}$, and 15.8\% in $\mathcal{F}_{\beta}$.

\begin{table}[h]
\centering
\caption{Ablation study for the effectiveness of loss function on MCOD benchmark dataset. $\uparrow/\downarrow$ denotes the larger/smaller is better. Best results are marked in \textbf{bold}.}
\renewcommand{\arraystretch}{1} 
\resizebox{\columnwidth}{!}{
\setlength{\tabcolsep}{7pt}
\begin{tabular}{c c c | c c c c}
\toprule
$\mathcal{L}^1_{seg}$ & $\mathcal{L}^2_{seg}$ &$\mathcal{L}^3_{seg}$ & $\mathcal{E}_{\xi} \uparrow$ & $\mathcal{S}_{m} \uparrow$  & $\mathcal{F}_{\beta} \uparrow$  & $\mathcal{M} \downarrow$ \\ \midrule
\checkmark & \cross & \cross & 0.888 & 0.858 & 0.799 & 0.003 \\
\cross & \checkmark & \cross & 0.250 & 0.278  & 0.007  & 0.517 \\
\cross & \cross & \checkmark & 0.254 & 0.312  & 0.007  & 0.433 \\
\rowcolor{wacvblue!10}
\checkmark & \checkmark & \checkmark & \textbf{0.972} & \textbf{0.881} & \textbf{0.805} & \textbf{0.002} \\ \bottomrule
\end{tabular}
}
\label{tab: ablation}
\end{table}

\subsubsection{Effectiveness of Loss Components.} 
In this section, we analyze the effectiveness of each loss component by selectively enabling the segmentation losses at the three prediction scales, where $\mathcal{L}^1_{seg}$ supervises the finest, full-resolution output $\hat{m}_1$ and $\mathcal{L}^2_{seg}$, $\mathcal{L}^3_{seg}$ supervises progressively coarser intermediate maps; the results are reported in Table~\ref{tab: ablation}. We observe that utilizing intermediate losses $\mathcal{L}^2_{seg}$ or $\mathcal{L}^3_{seg}$ in isolation leads to model collapse, as the final output layer lacks direct supervision. $\mathcal{L}^2_{seg}$ alone yields $\mathcal{F}_{\beta}$=0.007 with $\mathcal{M}$=0.517, and $\mathcal{L}^3_{seg}$ alone yields $\mathcal{F}_{\beta}$=0.007 with $\mathcal{M}$=0.433. In contrast, the primary loss $\mathcal{L}^1_{seg}$ alone yields competitive results, since it directly constrains the full-resolution output. However, it fails to capture fine-grained details. The full objective significantly improves performance by 9.5\% in $\mathcal{E}_{\xi}$ and 2.7\% in $\mathcal{S}_{m}$ over $\mathcal{L}^1_{seg}$ alone. This confirms that deep supervision facilitates robust gradient flow and multi-scale feature refinement, both of which are critical for delineating camouflaged boundaries.

\section{Conclusion}
\label{sec:conclusion}
In this paper, we introduced \textit{MSFormer}, an end-to-end transformer-based approach for multispectral camouflaged object detection. We propose an efficient strategy for incorporating multispectral inputs into a pyramid vision transformer, a cascaded decoder module that integrates hierarchical features from the encoder and produces effective prediction maps at multiple scales. Our proposed approach performs significantly well across different COD benchmarks for RGB, multispectral, and hyperspectral bands. In our future work, we plan to extend our proposed approach to multispectral videos as well.

{
    \small
    \bibliographystyle{ieeenat_fullname}
    \bibliography{refs}
}

\end{document}